\documentclass[sigconf]{acmart}

\AtBeginDocument{%
  }

\setcopyright{acmlicensed}
\copyrightyear{2025}
\acmYear{2025}
\acmDOI{XXXXXXX.XXXXXXX}

\acmConference[ACM XXXXX]{ACM Conference XX XXXXXXXX, XXXXXXXXXXXXXX, XXX XXXXXXXXXXXX}{XXXX XX--XX,
  2025}{XXXXXX, XXXXXX}

\acmISBN{XXX-X-XXXX-XXXX-X/XX/XX}

\usepackage{xcolor}
\usepackage{hyperref}
\usepackage{subfig}
\usepackage{graphicx}
\usepackage{multirow}
\usepackage{tcolorbox}

\newcommand{\new}[1] {{\color{black} #1}}

\begin{document}

\title{Neglected Risks: The Disturbing Reality of Children's Images in Datasets and the Urgent Call for Accountability}

\author{Carlos~Caetano}
\affiliation{%
  \institution{Instituto de Computação, Universidade Estadual de Campinas (UNICAMP)}
  \city{Campinas}
  \country{Brazil}
  }

\author{Gabriel~O.~dos~Santos}
\affiliation{%
  \institution{Instituto de Computação, Universidade Estadual de Campinas (UNICAMP)}
  \city{Campinas}
  \country{Brazil}
  }

\author{Caio~Petrucci}
\affiliation{%
  \institution{Instituto de Computação, Universidade Estadual de Campinas (UNICAMP)}
  \city{Campinas}
  \country{Brazil}
  }

\author{Artur~Barros}
\affiliation{%
  \institution{Instituto de Computação, Universidade Estadual de Campinas (UNICAMP)}
  \city{Campinas}
  \country{Brazil}
  }

\author{Camila~La\-ranjeira}
\affiliation{%
  \institution{Departamento de Ciência da Computação, Universidade Federal de Minas Gerais (UFMG)}
  \city{Belo Horizonte}
  \country{Brazil}
  }

\author{Leo~S.~F.~Ribeiro}
\affiliation{%
  \institution{Instituto de Ciências Matemáticas e de Computação, Universidade de São Paulo (USP)}
  \city{São Carlos}
  \country{Brazil}
}

\author{Júlia~F.~de~Mendonça}
\affiliation{%
  \institution{Instituto Alana}
  \city{Salvador}
  \country{Brazil}
}

\author{Jefersson~A.~dos~Santos}
\affiliation{%
  \institution{School of Computer Science, University of Sheffield}
  \city{Sheffield}
  \country{United Kingdom}
  }

\author{Sandra~Avila}
\affiliation{%
  \institution{Instituto de Computação, Universidade Estadual de Campinas (UNICAMP)}
  \city{Campinas}
  \country{Brazil}
  }

\renewcommand{\shortauthors}{Caetano et al.}

\newcommand{\todo}[1] {{\color{red}\textbf{[ToDo: #1]}}}

\begin{abstract}
Including children's images in datasets has raised ethical concerns, particularly regarding privacy, consent, data protection, and accountability. These datasets, often built by scraping publicly available images from the Internet, can expose children to risks such as exploitation, profiling, and tracking. Despite the growing recognition of these issues, approaches for addressing them remain limited. We explore the ethical implications of using children's images in AI datasets and propose a pipeline to detect and remove such images. As a use case, we built the pipeline on a Vision-Language Model under the Visual Question Answering task and tested it on the \textit{\#PraCegoVer} dataset. We also evaluate the pipeline on a subset of 100,000 images from the Open Images V7 dataset to assess its effectiveness in detecting and removing images of children. The pipeline serves as a baseline for future research, providing a starting point for more comprehensive tools and methodologies. \new{While we leverage existing models trained on potentially problematic data, our goal is to expose and address this issue. We do not advocate for training or deploying such models, but instead call for urgent community reflection and action to protect children's rights.} Ultimately, we aim to encourage the research community to exercise --- more than an additional --- care in creating new datasets and to inspire the development of tools to protect the fundamental rights of vulnerable groups, particularly~children.
\end{abstract}



\maketitle

\section{Introduction}

In recent years, the rapid advancement of artificial intelligence (AI) has raised important ethical concerns, particularly around privacy issues, consent, and accountability~\cite{Birhane:2021, Birhane:2024, Leu:2024:FAccT}. As multimodal AI systems evolve, increasingly demanding larger datasets to enhance their capabilities, worries regarding the inclusion of images of children in these training datasets emerge~\cite{Laranjeira:2022:FAccT, Yu:SP:2025, Diresta:2024, Arango:2023}. These datasets are typically built by scraping publicly available images from the Internet without the explicit consent of the individuals who own the data. This practice raises significant questions about individuals' awareness and control over their digital data, and it lacks transparency in how these images are subsequently employed in AI-based models. This is particularly concerning when it involves children because they constitute a group that is vulnerable to different forms of abuse and exploitation.

A prime example of these issues is the LAION family datasets~\cite{Schuhmann:2021:Laion400M, Webster:2023:deduplicationlaion2b, Schuhmann:2022:Laion5B}, large-scale image collections that have been widely used to train generative AI tools~\cite{Rombach:2022}. These datasets have been created by scraping images from the web without retaining a full repository of the images themselves. Instead, LAION maintains a list of URLs and metadata associated with the images, which makes it difficult to track or manage the content within the dataset. This is concerning not only from the privacy point of view but also due to the nature of the material included in these datasets. \text{\citet{Birhane:2021}} showed that LAION-400M contains image-text pairs related to rape, pornography, and racial and ethnic slurs. Furthermore, comparative audits~\cite{Birhane:2024, Birhane:2024:FAccT} of LAION-400M~\cite{Schuhmann:2021:Laion400M} and LAION-2B-en~\cite{Webster:2023:deduplicationlaion2b}\footnote{The LAION-2B-en is an English caption subset of LAION-5B~\cite{Schuhmann:2022:Laion5B}.} have demonstrated scaling the dataset amplified the presence of harmful and hateful content, including targeted and aggressive speech. Problematic language and imagery related to hate speech remain even after filtering for Not Safe For Work (NSFW)\footnote{NSFW is an internet acronym warning that content may be inappropriate in certain settings due to sexual, violent, or offensive material.} content, suggesting that existing filtering approaches are insufficient to eliminate harmful content. This raises important questions about the potential downstream harms that such datasets can perpetuate in AI systems, such as reinforcing harmful stereotypes or propagating violence~\cite{Birhane:2021} and highlights the need for more rigorous filtering and oversight in the creation of these datasets, mainly when dealing with sensitive content like images of children.

Human Rights Watch reported that a tiny-scale analysis of the LAION-5B dataset~\cite{Schuhmann:2022:Laion5B} revealed images of children, some of whom were easily identifiable through metadata, captions, and even URLs linked to the images. They reviewed less than 0.0001\% of the dataset's 5.85 billion images and captions, yet they uncovered disturbing instances of identifiable children. In one of the cases\footnote{\url{https://www.hrw.org/news/2024/06/10/brazil-childrens-personal-photos-misused-power-ai-tools}}, they found 170 photos of children from at least 10 states in Brazil. Some images included the children's names in the accompanying caption or the URL where the image was stored, making their identities easily traceable. Many images also provided details on when and where the photos were taken, compromising the children's privacy. Similarly, another analysis\footnote{\url{https://www.hrw.org/news/2024/07/03/australia-childrens-personal-photos-misused-power-ai-tools}} uncovered 190 photos of children from all of Australia's states and territories. Again, in some images, the children could be easily identifiable by the names included in the caption or URLs.

These findings are deeply concerning for several reasons. First, including images with identifiable people in datasets without their consent violates fundamental privacy rights. This is particularly troubling for children, who are highly vulnerable to various forms of exploitation, including sexual and commercial exploitation.  The unauthorized use of their personal information and images can lead to harmful consequences, such as tracking, profiling, and other malicious activities. Furthermore, as AI systems grow more sophisticated, the ability to generate highly realistic images of individuals — both children and adults — poses even greater risks to privacy and human dignity~\cite{Thiel:2023:Stanford, Shoaib:2023:ICCA}. The ease with which these children's images were identifiable, even in a tiny sample of the LAION-5B dataset, highlights the urgent need for stricter oversight, transparency, and accountability in creating and using datasets for AI~training. 

In light of these concerns, this paper investigates the detection of children\footnote{We adopt the United Nations (UN) definition of a child as anyone under the age of 18.} in images by using a Vision-Language Model (VLM) under Visual Question Answering (VQA) task. Our approach emphasizes the importance of balancing technical rigor with ethical responsibility, particularly in scenarios where such images may exist without consent or proper oversight. Through this work, we seek to provide a robust framework for addressing these challenges while fostering broader discussions on safeguarding children's privacy and rights in AI research. 

In summary, the contributions of this paper are fourfold:

\begin{enumerate}
    \item \textbf{A Pipeline for Removing Children's Images from Data\-sets:} We propose a VLM-based pipeline specifically designed to detect children's images and remove them from datasets, emphasizing high Recall to ensure comprehensive detection.
    
    \item \textbf{Evaluation Under Visual Question Answering (VQA) Methodology:} We experiment with eight model variations from multiple language decoders (including Vicuna~\cite{Chiang:2023:Vicuna}, Mistral~\cite{Jiang:2023:Mistral}, and Hermes-Yi~\cite{NousHermesYi:2024}) to asses their effectiveness in detecting  children's images.

    \item \textbf{Benchmark Analysis:} We apply our pipeline to the \textit{\#PraCegoVer} dataset~\cite{Santos:2022:PraCegoVer} (272,245 images) and meticulously curate a subset of 1,364 images from \textit{\#PraCegoVer} by manually selecting images containing identifiable children in different contexts. We also apply our pipeline to a subset of 100,000 images from the Open Images V7 dataset~\cite{Benenson:2022:OpenImagesV7}, uncovering critical challenges such as annotation inconsistencies.

    \item \textbf{Promotion of Ethical AI Practices:} Beyond providing a practical tool for removing child images from datasets, we aim to spark an essential and overdue discussion about the ethical implications of having child images in large-scale AI datasets. This includes images taken without proper consent or adherence to ethical guidelines, raising concerns about privacy violations and potential misuse. By addressing this gap, we encourage researchers to adopt more responsible dataset curation practices. We hope this work sparks discussions on the ethical implications and risks associated with the inclusion of children's data in large-scale AI training~datasets.
\end{enumerate}

This paper is structured as follows. In Section~\ref{sec:background-related-work}, we provide an overview of the background and related work, discussing ethical concerns in dataset creation and efforts to mitigate risks. In Section~\ref{sec:methodology}, we detail the proposed \text{VLM-based} pipeline, including model configurations and dataset considerations. In Section~\ref{sec:experimental-analysis}, we present the experimental setup and results, followed by an analysis of the findings on the evaluated datasets. The limitations of our work are highlighted in Section~\ref{sec:limitations}. Finally, in Section~\ref{sec:conclusion}, we conclude the paper with a summary of our contributions and suggestions for further research. Ethical considerations and adverse impacts statements are presented to ensure transparency and responsibility. 
\section{Background and Related Work}\label{sec:background-related-work}

\paragraph{Ethical Concerns in Dataset Creation.}
Web scraping for dataset creation raises concerns about privacy and consent, as collecting images without explicit permission violates fundamental rights to privacy and control over personal data~\cite{Rupp:2024:CLSR, Xiongbiao:2024:TP, Mancosu:2020:SMS}. Including identifiable details like names, locations, and timestamps heightens risks of misuse, particularly for children, who face greater vulnerability and lasting impacts from privacy breaches.

The LAION-5B dataset~\cite{Schuhmann:2022:Laion5B}, widely used in training generative AI models, has been criticized for lacking transparency and oversight. Human Rights Watch's report revealed it contains images of children that could be identified through sensitive metadata. Similarly, \citet{Birhane:2021:WACV} found inappropriate content in ImageNet-ILSVRC-2012, raising concerns about privacy and consent. These examples highlight the urgent need for stricter regulations and ethical guidelines in dataset~creation.

Additionally, \citet{Yu:SP:2025} explored generative AI use among teenagers, highlighting risks and the gap between parents' and children's safety perceptions, stressing the need for safety mechanisms and parental controls. Similarly, \citet{Diresta:2024} emphasized how malicious actors exploit AI-generated images, reinforcing the importance of oversight to protect vulnerable groups, including children. \citet{Arango:2023} extended this to societal impacts, showing that ethical transparency in using AI-generated versus real images is crucial for maintaining public trust in sensitive applications like charity campaigns.

Regarding web scraping, different studies~\cite{Brown:2024:arXiv, Mancosu:2020:SMS, Jayasankar:2023:SSRN} have investigated the ethical challenges around it. \text{\citet{Brown:2024:arXiv}} propose a framework to guide researchers through the legal, ethical, institutional, and scientific considerations of web scraping for research purposes. It emphasizes the importance of institutional oversight, such as Institutional Review Boards, in ensuring ethical practices.  \citet{Mancosu:2020:SMS} highlight ethical and legal issues in scraping social media data, advocating pseudonymization to protect privacy. \citet{Jayasankar:2023:SSRN} explore the ethical landscape of data scraping specifically for generative AI, emphasizing privacy concerns, consent, and the need for diverse and representative datasets to avoid bias.

\paragraph{Mitigating Risks.} 

Efforts to mitigate the risks associated with including sensitive data in AI datasets have been limited and largely reactive. Some initiatives use filtering mechanisms, such as detection and recognition algorithms, to exclude sensitive content, but these methods often lack precision for addressing challenges with children's images. NSFW classifiers, commonly used to remove explicit content, are prone to biases. \citet{Leu:2024:FAccT} show these classifiers disproportionately misclassify images of women, people with lighter skin tones, and younger individuals, raising fairness concerns. This is particularly problematic for children's images, where classifiers may struggle to differentiate harmless from inappropriate content based on age and demographics.

Studies have explored ethical challenges in curating datasets with children's images. \citet{AlAzani:2022:Access} review children's databases, addressing issues like consent acquisition and ethical oversight, stressing the need for rigorous ethical standards. Similarly, \citet{Moore:2019:PR} review the use of machine learning in pediatric contexts, discussing technical and ethical challenges that call for targeted solutions. Additionally, \citet{Florindi:2024:ICCSA} examine broader ethical concerns, including the misuse of AI tools to create harmful content like virtual child sexual abuse material. The ethical dilemmas outlined in their work provide a foundation for understanding the importance of responsible data curation and its impact on AI applications.  

Adopting data curation practices from fields like archives and data stewardship offers a pathway to more ethical dataset creation. \citet{Bhardwaj:2024:FAccT} propose a framework for evaluating machine learning datasets through the lens of data curation. Their work underscores the difficulty researchers face in applying standard data curation principles to machine learning datasets, particularly when it comes to ensuring comprehensive and responsible documentation of dataset creation processes. This framework provides valuable guidance for ensuring the ethical curation of AI training datasets, including those involving images of children.

Mitigating the risks associated with sensitive content in AI training datasets aimed at detecting and eliminating explicit material. In a technical report by \citet{Thiel:2023:StanfordCSAMIdentifying}, they found child sexual abuse material (CSAM) in the LAION-5B dataset. They employed advanced methodologies --- including hash matching, \text{k-nearest} neighbors queries, and machine learning classifiers --- to identify hundreds of instances of known CSAM and verify new candidates with external parties. This highlights a critical gap in existing dataset curation practices and underscores the need for proactive measures to eliminate harmful content during the dataset construction phase.

While \citet{Thiel:2023:StanfordCSAMIdentifying} focus on detecting CSAM from LAION-5B dataset, our work addresses a broader and distinct concern related to the inclusion of images of children in large-scale datasets constructed without consent or through unethical practices. We aim to \textit{detect and remove} all images of children — regardless of their content — due to privacy, consent, and ethical concerns. Our work emphasizes the ethical and legal implications of including images of children without consent, as well as the risks associated with their potential misuse in AI systems.
\section{Methodology}\label{sec:methodology}

We aim to assess the effectiveness and reliability of our vision-language-based pipeline, designed to detect and remove images of children from datasets. The pipeline employs a VLM under the VQA task to create an automated approach for dataset filtering.

We also consider the ethical implications of the pipeline's design and operation, ensuring that it minimizes biases and avoids the exclusion of non-child images. By establishing a comprehensive and systematic methodology, this study aims to provide a replicable framework for researchers and practitioners to adopt in addressing similar ethical challenges in AI dataset creation and usage.

\subsection{Pipeline Design}
\label{subsec:pipeline_design}

\begin{table*}[t]
	\centering
    	\caption{Comparison of prompt variations and response examples. The table presents the four evaluated prompt variations (indexed from 0 to 3), along with example outputs generated by the model for a given input image. The responses illustrate the impact of prompt design on the verbosity and clarity of the model's answers, highlighting the progression from verbose descriptions to concise binary classifications. Image from \textit{\#PraCegoVer}~\cite{Santos:2022:PraCegoVer} dataset.}
		\label{tab:prompts}
		\begin{tabular}{cccclc}
			\toprule
			\textbf{Input Image} && \textbf{\#} && \textbf{Prompt} & \textbf{Output Response}\\
			\toprule
		      \multirow{14}{*}{\includegraphics[width=0.2\textwidth, height=0.2\textwidth]{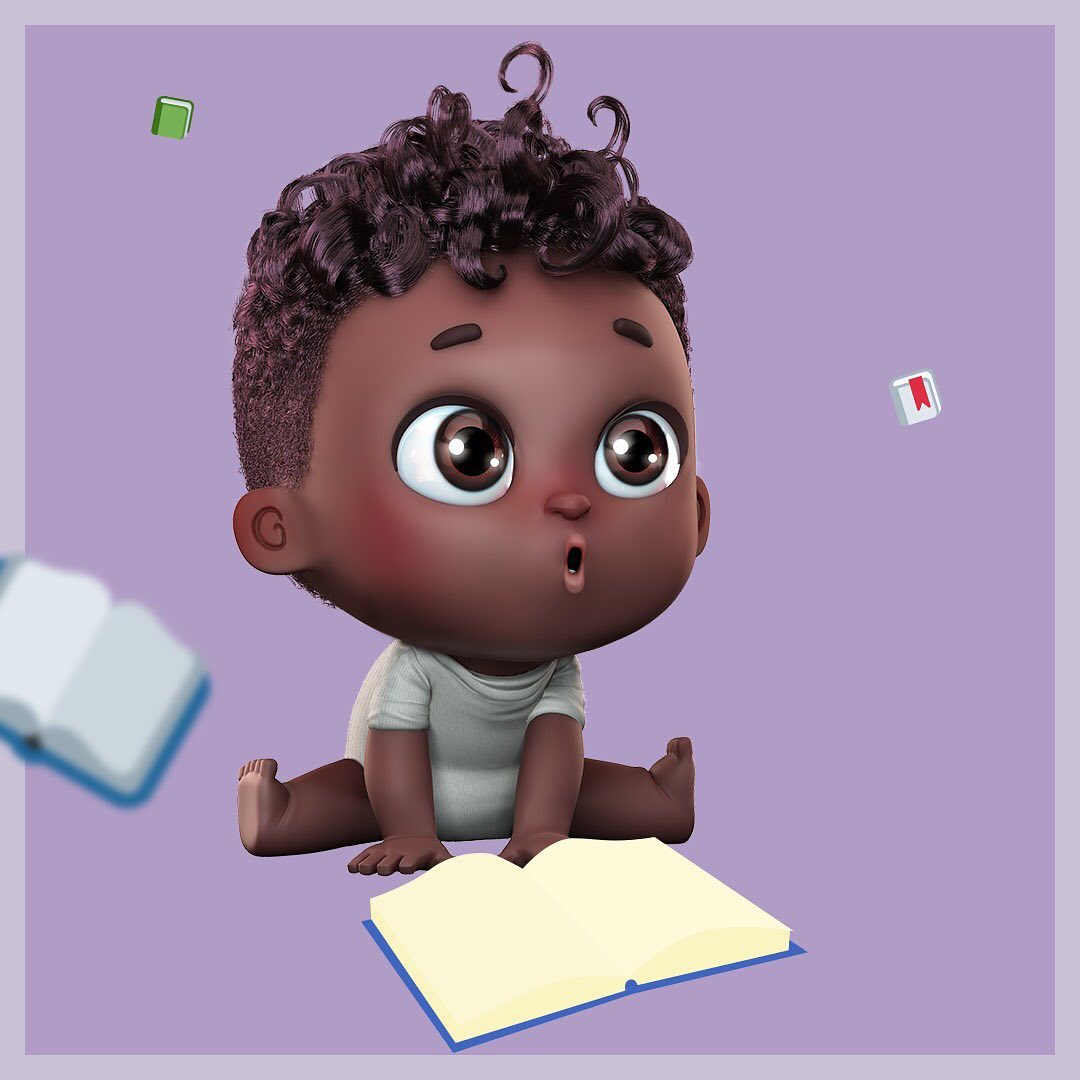}}
                && \multirow{3}{*}{0} && \multicolumn{1}{p{6.2cm}|}{\multirow{3}{*}{Are there any children in the picture?}} & \multicolumn{1}{p{2.88cm}}{Yes, there is a child in the picture, specifically a baby or a toddler.} \\
                \cmidrule{2-6}
                && \multirow{2}{*}{1} && \multicolumn{1}{p{6.2cm}|}{Are there any children in the picture? Answer with only ``Yes'' or ``No''.}
 & \multirow{2}{*}{Yes} \\
                \cmidrule{2-6}
                && \multirow{3}{*}{2} && \multicolumn{1}{p{6.2cm}|}{Are there any children in the picture? Disconsider any cartoons or digital illustrations, consider only real children. Answer with only ``Yes'' or ``No''.}
 & \multirow{3}{*}{No} \\
                \cmidrule{2-6}
                && \multirow{5}{*}{3} && \multicolumn{1}{p{6.2cm}|}{Are there any children in the picture? Disconsider children facing away or with the face not visible. Also, disconsider any cartoons or digital illustrations, consider only real children. Answer with only ``Yes'' or ``No''.}
 &  \multirow{5}{*}{No} \\
                
			\bottomrule
		\end{tabular}
\end{table*}

We approach child detection as a VQA task and leverage VLMs' capability to handle complex tasks. The pipeline operates by first receiving a collection of images that may include both child and non-child content. Each image is processed individually, and a carefully designed textual prompt is provided to the model. \autoref{tab:prompts} presents the four variations of prompts we considered. \new{We adopted a cumulative approach to balance ethical considerations with technical robustness. Moreover, the prompts are presented cumulatively to illustrate the iterative refinement process. This design choice was crucial for ensuring scalability and minimizing computational overhead, as it avoids the inefficiencies associated with cascade prompting systems (i.e., the sequential use of multiple prompts).} For preliminary tests, we used the first variation (prompt \#0 from \autoref{tab:prompts}). However, the responses from this prompt were excessively verbose, complicating the downstream parsing required for binary classification. To address this, we adapted the prompt to explicitly instruct the model to answer with a binary response of ``Yes'' or ``No'' (prompt \#1 from \autoref{tab:prompts}).

While prompt \#1 improved the clarity and parsability of responses, it was necessary to introduce additional variations to account for ethical considerations and the nuanced nature of detecting children in images. To that end, we introduced prompt \#2 to exclude non-human representations, such as cartoons or digital illustrations, ensuring that only real children were considered in the classification. 

We achieved further refinement with prompt \#3. We designed this prompt to not consider images where children are not identifiable due to occlusion or lack of visible facial features. Such images are acceptable to keep in the dataset as they do not pose significant privacy or ethical concerns or lead to potential risks such as tracking or profiling. By excluding unnecessary classifications of these images as child content, prompt \#3 helps ensure that the dataset retains as much useful data as possible without compromising ethical considerations. These progressive refinements illustrate the importance of precise prompts designed to provide both the evaluation pipeline's technical robustness and ethical~compliance. By employing the four prompt variations, the study aims to comprehensively assess the interplay between model structure, scale, and prompt effectiveness in the context of detecting children in~images.

\begin{figure*}[t]
   \centering
   \includegraphics[width=\linewidth]{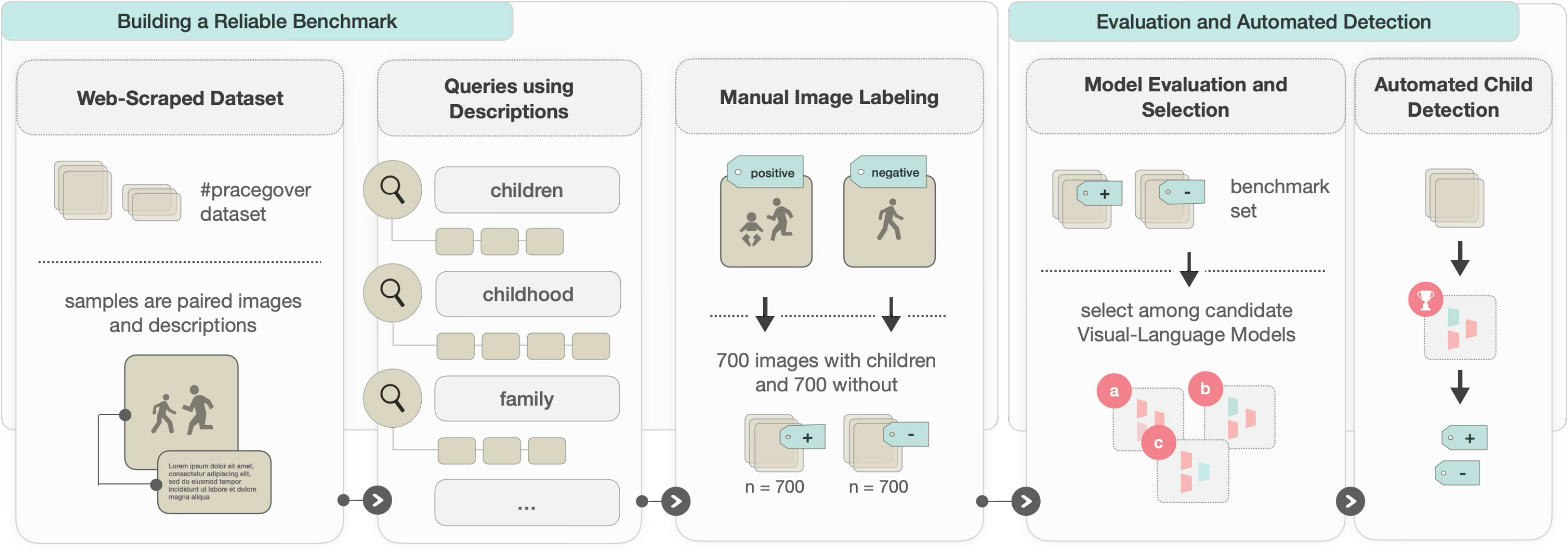}
   \caption{The methodology overview is divided into two main steps. (i)~\emph{Building a Reliable Benchmark} illustrates the steps employed to curate a subset of the \textit{\#PraCegoVer} dataset, leveraging metadata to ensure reliable annotations for distinguishing children from adults. (ii)~\emph{Evaluation and Automated Detection} details the pipeline based on VLMs, designed to assess whether an image contains a child.}
   \Description{The methodology overview is divided into two main steps. (i)~\emph{Building a Reliable Benchmark} illustrates the steps employed to curate a subset of the \textit{\#PraCegoVer} dataset, leveraging metadata to ensure reliable annotations for distinguishing children from adults. (ii)~\emph{Evaluation and Automated Detection} details the pipeline based on VLMs, designed to assess whether an image contains a child.}
   \label{fig:vqa_pipeline}
\end{figure*}

After obtaining the binary responses from the model (i.e., ``Yes'' or ``No''), the pipeline parses the outputs to classify the images. The classification determines whether the model has correctly detected the presence of children in the images, allowing for detailed performance analysis and insights into the strengths and limitations of the approach. \autoref{fig:vqa_pipeline} illustrates the employed pipeline as~\emph{Evaluation and Automated Detection}.

\subsubsection{Underlying Vision-Language Models}

The models used in this pipeline are built upon the foundations of multimodal instruction tuning. These models connect a vision encoder, such as CLIP, with a language decoder (e.g., Vicuna), enabling end-to-end training on vision-language data. By fine-tuning on instructional datasets, these models demonstrate robust general-purpose visual and language understanding capabilities. To that end, we employed the Large Language and Vision Assistant (LLaVA)~\cite{Liu:2023:LLaVA}. \new{LLaVA's fine-grained instruction tuning enables precise and interpretable responses, making it suitable for our target task.}

The adoption of VLMs in this pipeline is motivated by their high accuracy on benchmark multimodal reasoning tasks and their ability to generate precise, instruction-following responses. The fine-grained multimodal instruction tuning further ensures that the models can effectively handle diverse and challenging visual~inputs.

We evaluate three distinct language decoder models with LLaVA to investigate the impact of model architecture and prompt design on the pipeline’s performance, including Vicuna~\cite{Chiang:2023:Vicuna} (six models), Mistral~\cite{Jiang:2023:Mistral} (one model), and Hermes-Yi~\cite{NousHermesYi:2024} (one model), each bringing unique characteristics to the multimodal reasoning \text{process}. 

\subsubsection{Advantages of the VLM-Based Approach}

Adopting VLMs for detecting images containing children brings several benefits. One primary advantage is that, being trained on large datasets, these models can adapt to a wide range of visual contexts. Such adaptability enables them to precisely detect children in images, even under challenging conditions such as different lighting, varied backgrounds, or unusual poses. Then, this approach is highly robust for analyzing diverse~datasets.

Another benefit of a VLM-based approach lies in its ability to respond directly to natural language prompts. Unlike traditional image classification methods that rely on predefined labels and static feature sets, VLMs can tackle nuanced queries, such as identifying the presence of children in an image, without requiring extensive feature engineering or complex heuristic rules. By using straightforward prompts, such as ``Are there any children in this image?'', these systems remain both interpretable and easy to~use.

Additionally, the VLM-based approach supports scalability and extensibility. By decoupling the visual encoding from the text generation, the framework enables experimentation with different language decoders and visual encoders. This modular design not only allows for comparative evaluations across models but also simplifies the integration of future advancements in vision and language processing. Consequently, the pipeline remains flexible and capable of adapting to the rapid evolution of AI technologies.

Finally, this approach promotes transparency and accountability in processing datasets. Explicit prompts and interpretable outputs foster trust and allow researchers to trace and validate the decision-making process. Overall, the VLM-based approach combines adaptability, interpretability, and scalability, making it a powerful tool to mitigate the inclusion of children's images in datasets.

\subsection{Evaluation Protocol}

\subsubsection{\textit{\#PraCegoVer} Dataset}
\label{subsubsec:pracegover}
We adopt \textit{\#PraCegoVer}~\cite{Santos:2022:PraCegoVer} to validate our children detection and removal pipeline as a use case. This dataset is one of the largest proposed for vision-and-language tasks, featuring texts written by native Brazilian Portuguese speakers. It originally comprised over 530,000 samples, each containing an image paired with a caption. The dataset is built upon data collected from public profiles on Instagram. Thus, photos of children are expected in this dataset. \new{In addition to captions, the dataset also provides visual descriptions — manual textual descriptions detailing the image content (e.g., ``The image contains a child in front of a school. The child is a boy wearing a uniform composed of a shirt and shorts'').}

Given its data source, \textit{\#PraCegoVer} is considered a noisy dataset, as it relies on annotations provided by social media users. Therefore, before applying our pipeline, we perform a preprocessing step to clean the dataset by removing near-duplicate samples and instances where the image and text are unrelated. First, we remove near-duplicate samples using the method proposed by \citet{Santos:2022:PraCegoVer}. Then, to identify cases in which captions fail to describe the visual content of the images, we utilize CAPIVARA~\cite{Santos:2023:CAPIVARA}. This model is based on the CLIP~\cite{Radford:2021:CLIP} architecture, trained to align images with text, and it is specialized in Portuguese. The model computes a similarity score between 0 and 1 for a given image and caption pair. We process all deduplicated samples with CAPIVARA, passing each image-caption pair through the model, and we remove the examples whose similarity score is lower than a given threshold\footnote{We used a similarity threshold of $0.2$}. After this process, the final dataset consists of 272,245 samples.

\paragraph{Validation Subset}
After cleaning \textit{\#PraCegoVer}, we select and manually annotate a subset to evaluate our method. To construct this subset, first, we define a list of keywords in Portuguese related to ``children'',  ``childhood'', ``family'', ``school'', ``beach'', ``trip'', ``party'',  ``work'', and  ``university''. For each category, we randomly sample 500 instances whose text contains one of the keywords, resulting in 4500 examples. Next, we manually analyze the selected images and pick 701 images containing identifiable children in different contexts. To ensure a balanced evaluation, we also select a similar number (663) \new{of negative examples. Under the negative samples, we also selected} images specifically chosen to confound the models, featuring scenes and elements visually similar to those in the children's class. \new{The availability of captions and the visual descriptions was critical for defining and verifying whether an image contained a child. This rich metadata enabled us to resolve ambiguities, such as distinguishing teenagers from adults. For instance, in cases where the context suggested a teenager (e.g., an advertisement from a college account showing children in uniforms), we could confidently classify the individual as a child under the UN definition. Captions often provided contextual clues (e.g., ``my son’s first day at school'' or ``family trip with the kids''), while visual descriptions allowed us to assess the child’s visual characteristics, such as clothing, setting, and apparent age.} The textual information is only used to create this validation subset \new{and we emphasize that our final classification pipeline does not used textual information (i.e., only images are processed). While keyword-based sampling may introduce some bias, we believe the subset’s diversity mitigates this concern.} \autoref{fig:vqa_pipeline} illustrates the subset creation as~\emph{Building a Reliable~Benchmark}.

\subsubsection{Open Images Dataset}
\label{subsubsec:openimages}

To evaluate our method on a larger scale, we conducted additional experiments using a subset of the Open Images V7 dataset~\cite{Benenson:2022:OpenImagesV7}. Open Images is a large-scale dataset with various annotations, including image-level labels, detection boxes, and segmentation masks. Within the dataset's approximately 20k mapped concepts, 600 are designated as ``boxable'' detection categories, annotated across nearly 1.8 million samples. The structured hierarchy established for these classes includes categories that are differentiated by age and gender expression. Specifically, the classes \textit{woman}, \textit{man}, \textit{boy}, and \textit{girl} are organized hierarchically under \textit{person}. Although most datasets with age-related labels focus on faces, Open Images consists of unconstrained Flickr images.

Understanding the Open Images annotation process is crucial to interpret our results. According to \citet{kuznetsova2020open}, for image-level labels, annotators were presented with an image and a list of automatically extracted concepts, which could include redundancies (e.g., person, woman, and girl for an image of a single person). Annotators were asked to confirm whether each class was present. Any missed concepts not detected by the automatic extraction remain unknown. A class was considered positively identified if the majority of seven crowdsource annotators agreed on its presence or a Google internal annotator labeled it as present. From image-level labels, detection ground-truths were acquired by instructing a Google internal team to assign bounding boxes for all concepts deemed present. The authors attempted to avoid redundancy by removing a class from the detection annotation stage if another class hierarchically subordinate to it was also present in the same image. For instance, if the classes person, woman, and girl were identified, only the person class was excluded.

\paragraph{Validation Subset}
We selected a subset of 100,000 samples from Open Images to create a balanced set of images. \new{We relied on the provided age-related annotations (e.g., ``boy'', ``man'', ``girl'', ``woman'') to determine whether a child was present in the image.} This subset consisted of 50,000 samples with detection labels for boys or girls representing images of children and an additional 50,000 sample in which classes related to children were absent but either a woman or a man was present. By leveraging detection labels, we were able to exclude targets labeled as ``depiction,'' ensuring that images of children only contained real people.

\section{Experimental Analysis}\label{sec:experimental-analysis}

In this section, we present the experiment results and analysis conducted using the LLaVA to perform VQA. The objective was to evaluate it as a binary classification system determining whether a given image contains a child. The input to the model was an image accompanied by a textual prompt designed to elicit a binary response. Four distinct prompts were initially designed, as shown in \autoref{tab:prompts}. The prompt \#0 was exploratory but yielded verbose answers that complicated the parsing and classification process. Given that, the evaluation focuses on the performance of models when employing prompts \#1, \#2, and \#3. We designed these prompts to refine the classification criteria progressively, ensuring alignment with ethical considerations while avoiding unnecessary exclusion of valid dataset content.

By restricting our analysis to these three prompts, we concentrate on the optimized configurations that ensure binary responses and address privacy-related constraints, such as avoiding misclassifying images with non-identifiable children or illustrations. The experiments aim to assess the interplay between model architecture, prompt design, and classification metrics.

The experimental pipeline was executed using distinct model configurations, encompassing six Vicuna-based models, one Mistral-based model, and one Hermes-Yi-based model. These models include variations in size and versions, resulting in a total of eight distinct configurations. Each model was evaluated under three operational categories --- complex reasoning, conversational context, and detailed descriptions --- resulting in 24 model combinations for each~prompt.

All experiments were performed with two NVIDIA A100 GPUs, ensuring sufficient computational resources for processing large-scale datasets. We evaluated the aforementioned combinations on a subset of the \textit{\#PraCegoVer} dataset, previously annotated and curated as described in the third-level Section~\ref{subsubsec:pracegover}. We performed additional experiments on the Open Images V7 subset but considered only the model with the best results on \textit{\#PraCegoVer} subset.

\subsection{Results on \textit{\#PraCegoVer} Subset}

The evaluation on the subset of the \textit{\#PraCegoVer} dataset focused on the metrics of Recall and False Positive Rate (FPR), given their critical importance to the study's ethical objectives. Recall directly measures the model's ability to detect images containing children, ensuring such instances are not overlooked. FPR quantifies the misclassification of non-child images, reflecting the model's specificity. These metrics provide a balanced view of the performance trade-offs.

The evaluation of our pipeline revealed distinct performance patterns across different model configurations, categories, and prompts. \autoref{tab:pracegover_subset} summarizes the Recall and FPR values for each configuration, facilitating a comparative analysis.

\begin{table*}[!t]
	\centering
\caption{Performance comparison of multimodal models across categories and prompt variations on the subset of the \textit{\#PraCegoVer} dataset \cite{Santos:2022:PraCegoVer}. The table presents the Recall and False Positive Rate (FPR) achieved by each model configuration, categorized by the model's architecture, prompt variation, and response category (\textbf{complex} reasoning, \textbf{conv}ersational context, and \textbf{detail}ed~descriptions).}
		\label{tab:pracegover_subset}
		\begin{tabular}{lccc|cc|cc}
			\toprule
             &  & \multicolumn{2}{c}{\textbf{prompt \#1}} & \multicolumn{2}{|c|}{\textbf{prompt \#2}} & \multicolumn{2}{c}{\textbf{prompt \#3}}\\
			\textbf{Model} & \textbf{Category} & \textbf{Recall (\%)} & \textbf{FPR (\%)} & \textbf{Recall (\%)} & \textbf{FPR (\%)} & \textbf{Recall (\%)} & \textbf{FPR (\%)}\\
			\toprule
                \multirow{3}{*}{llava-v1.5-7b~\cite{Liu:2024:CVPR:LLaVA1.5}} & complex & 97.7 & 24.3 & 96.7 & 20.2 & 98.4 & 32.3 \\
                & conv & 97.7 & 23.1 & 97.1 & 21.6 & 98.4 & 30.6 \\
                & detail & 98.0 & 22.6 & 97.3 & 19.9 & 98.6 & 30.8 \\
                \midrule
                \multirow{3}{*}{llava-v1.5-7b-lora~\cite{Liu:2024:CVPR:LLaVA1.5}} & complex & 97.1 & 19.6 & 96.4 & 15.4 & 96.6 & 19.0 \\
                & conv & 96.9 & 20.2 & 95.9 & 14.9 & 96.9 & 18.7 \\
                & detail & 97.0 & 19.3 & 96.3 & 15.4 & 97.0 & 18.3 \\
                \midrule
                \multirow{3}{*}{llava-v1.5-13b~\cite{Liu:2024:CVPR:LLaVA1.5}} & complex & 98.0 & 21.0 & 98.3 & 21.3 & 98.9 & 22.3 \\
                & conv & 98.1 & 21.3 & 98.3 & 22.0 & 98.9 & 22.9 \\
                & detail & 98.4 & 20.8 & \textbf{98.9} & 22.3 & 98.7 & 22.6 \\
                \midrule
                \multirow{3}{*}{llava-v1.5-13b-lora~\cite{Liu:2024:CVPR:LLaVA1.5}} & complex & 98.0 & 22.2 & 98.0 & 17.8 & 97.3 & 14.8 \\
                & conv & 97.9 & 22.8 & 98.0 & 17.6 & 97.4 & 15.4 \\
                & detail & 98.3 & 22.0 & 97.3 & 16.9 & 97.6 & 14.6 \\
                \midrule
                \multirow{3}{*}{llava-v1.6-vicuna-7b~\cite{Liu:2024:LLaVA1.6-NeXT}} & complex & 98.0 & 17.3 & 98.4 & 20.4 & 98.7 & 25.3 \\
                & conv & 98.0 & 18.9 & 98.6 & 19.8 & 98.9 & 23.8 \\
                & detail & 98.3 & 17.6 & 98.3 & 18.9 & \textbf{99.0} & 24.4 \\
                \midrule
                \multirow{3}{*}{llava-v1.6-vicuna-13b~\cite{Liu:2024:LLaVA1.6-NeXT}} & complex & \textbf{98.7} & 18.4 & 96.9 & \textbf{12.4} & 95.4 & \textbf{13.3} \\
                & conv & 98.4 & 20.1 & 96.4 & 13.0 & 95.6 & 14.9 \\
                & detail & 98.6 & 19.0 & 96.4 & 12.7 & 96.0 & 13.7 \\
                \midrule
                \multirow{3}{*}{llava-v1.6-mistral-7b~\cite{Liu:2024:LLaVA1.6-NeXT}} & complex & 86.6 & 7.5 & 40.4 & 42.4 & 14.6 & 61.8 \\
                & conv & 87.0 & \textbf{6.2} & 39.4 & 45.4 & 14.7 & 62.1 \\
                & detail & 86.6 & 6.5 & 39.8 & 44.8 & 16.5 & 61.5 \\
                \midrule
			\multirow{3}{*}{llava-v1.6-34b~\cite{Liu:2024:LLaVA1.6-NeXT}} & complex & 97.6 & 18.3 & 97.7 & 33.9 & 95.3 & 52.6 \\
                & conv & 97.4 & 17.0 & 97.7 & 34.1 & 95.9 & 57.2 \\
                & detail & 98.0 & 17.6 & 97.9 & 33.6 & 95.4 & 54.8 \\	
                \bottomrule
		\end{tabular}
\end{table*}

Configurations using prompt \#1 demonstrated the capability to detect children in images with high Recall while maintaining relatively low FPR. The llava-v1.6-mistral-7b (category: conversation) achieved a Recall of 87.0\% with an FPR of 6.2\%, effectively minimizing false positives while maintaining substantial detection capabilities. On the other hand, the llava-v1.6-vicuna-13b (category: complex reasoning) reached a Recall of 98.7\%, ensuring that nearly all images containing children were detected, but with a higher FPR of 18.4\%. 

Prompt \#2 introduced refined criteria to exclude cartoons and digital illustrations. This specificity enhanced the pipeline’s focus on real child images while maintaining strong detection performance. The llava-v1.5-13b (category: detailed description) achieved a high Recall of 98.9\%, ensuring that nearly all child images were detected, with an FPR of 22.3\%. Similarly, the llava-v1.6-vicuna-13b (category: complex reasoning) achieved a Recall of 96.9\% with a reduced FPR of 12.4\%, minimizing misclassifications while upholding high detection rates. 

Prompt \#3 further refined the detection process by excluding images where children were not identifiable due to occlusion or lack of visible facial features. The llava-v1.6-vicuna-13b (category: complex reasoning) achieved a Recall of 95.4\% with a lower FPR of 13.3\%, demonstrating the effectiveness of Prompt \#3 in addressing ethical concerns while maintaining a robust detection framework. The llava-v1.6-vicuna-7b (category: detailed description) achieved the highest, and impressive, Recall of 99.0\%, ensuring that virtually all child images were detected, despite a higher FPR of~24.4\%. 

The results across all prompts underscore the importance of prioritizing Recall to ensure that all images containing children are detected and removed. Achieving high Recall aligns with the ethical objectives of protecting children’s privacy and mitigating risks associated with their inclusion in datasets. While lower FPR values are desirable for reducing misclassification of non-child images, the primary goal remains the comprehensive detection and removal of child images. In real-world applications, prioritizing Recall ensures that no child image goes undetected, reinforcing the ethical imperatives central to this study.

\new{Finally, we highlight that experiments with all the prompts were performed only on the \textit{\#PraCegoVer} subset to develop the final Prompt \#3, with the aim of applying it to full datasets. This strategy ensured prompt optimization on a smaller, controlled set before scaling, allowing for a single model query per image. This design choice was essential to maintain scalability and reduce computational overhead, avoiding the inefficiency of a cascade prompting system (i.e., using multiple prompts sequentially).}

\subsection{Results on  \textit{\#PraCegoVer} Dataset}

Since llava-v1.6-vicuna-7b (category: detailed description) in combination with prompt \#3 yielded the best results (i.e., with the highest Recall), we applied it to the entire \textit{\#PraCegoVer} dataset (272,245 images). The number of images classified as containing children is 41,900 (15.39\% of the dataset), while those classified as non-child images total 230,345 (84.61\%). These results highlight the presence of a significant proportion of images flagged as potentially sensitive, underscoring the need for further scrutiny and potential exclusion from datasets to ensure ethical compliance.

The emphasis on Recall in this broader dataset analysis reinforces the ethical rationale for prioritizing the identification of child-related content. Misclassifying such images as non-child content could result in severe consequences, including breaches of privacy and the retention of ethically non-compliant data.

\begin{figure*}[!t]
	\begin{tabular}{ccc}
    \centering
    \includegraphics[width=0.25\textwidth]{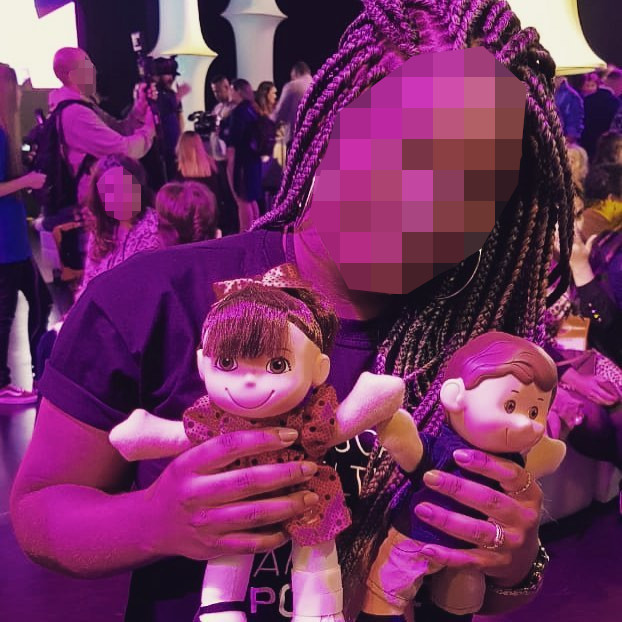} & 
    \includegraphics[width=0.25\textwidth]{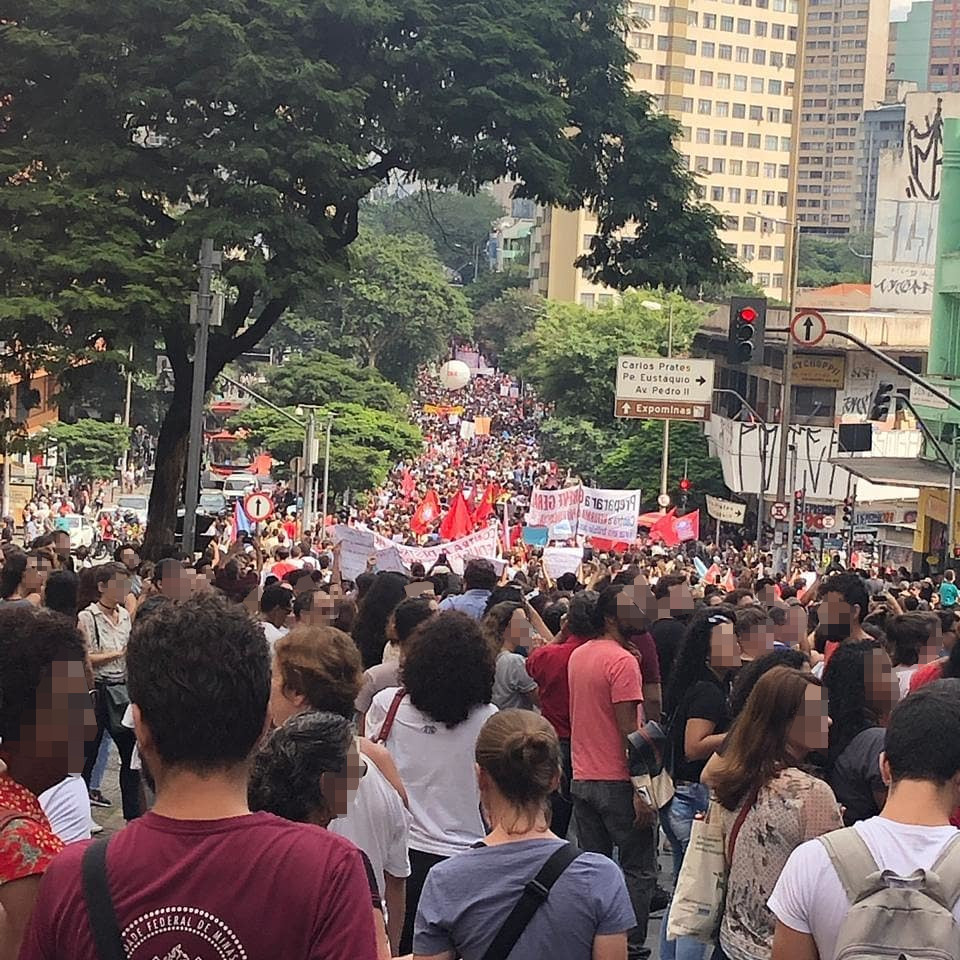} &   
    \includegraphics[width=0.25\textwidth]{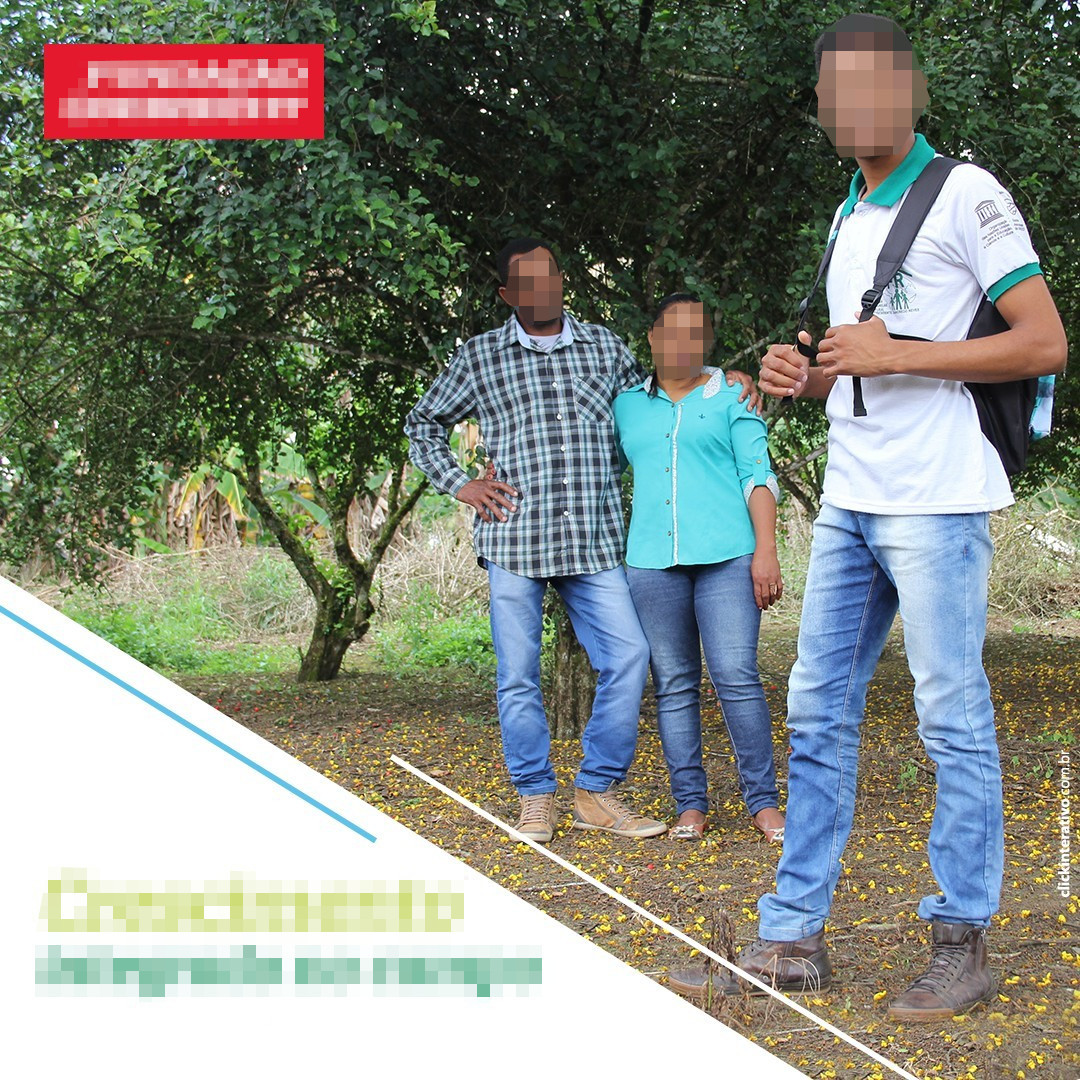} \\
    \footnotesize{(a) Contextual association of toys with children} & 
    \footnotesize{(b) Crowded scenes} & 
    \footnotesize{(c) Teenagers} \\
	\end{tabular}
	\caption{Examples of False Positive (FP) and False Negative (FN) classifications in the \textit{\#PraCegoVer} subset. The FP examples include (a)~a woman holding dolls, mistakenly identified as containing a child, and (b) a crowded scene with no children. The FN case (c)~depicts a teenager, misclassified as not containing a child.}
    \Description{Examples of False Positive (FP) and False Negative (FN) classifications in the \textit{\#PraCegoVer} subset. The FP examples include (a) a woman holding dolls, mistakenly identified as containing a child, and (b) a crowded scene with no children. The FN case (c) depicts a teenager, misclassified as not containing a child.}
	\label{fig:pracegover_FP_FN}
\end{figure*}

To further illustrate the challenges faced during classification, \autoref{fig:pracegover_FP_FN} provides representative examples of False Positive (FP) and False Negative (FN) classifications. For instance, FP cases include an adult woman holding two dolls, where the model likely misclassified the image due to the contextual association of dolls with children, and a crowded scene with numerous individuals but no children, where overlapping subjects and visual complexity led to an error. Conversely, the FN case features a teenagers, misclassified as adults due to their mature appearance. These examples underscore the model's struggles with contextual artifacts, crowded scenes, and the inherent difficulty of distinguishing teenagers from adults.

\subsection{Results on Open Images V7 Subset}

To assess the scalability of our approach, we performed additional experiments on a subset of the \text{Open Images V7} dataset~\cite{Benenson:2022:OpenImagesV7}. The subset's selections and creation process are detailed in the third-level in  Section~\ref{subsubsec:openimages}. For this experiment, we employed the configuration with the best-performing model on the \textit{\#PraCegoVer} subset --- llava-v1.6-vicuna-7b (category: detailed description) in combination with prompt \#3. The results revealed a recall of 71.7\% and a FPR of 16.2\%. 

While these results underscore the model's ability to detect child images, the lower Recall and higher FPR compared to the \textit{\#PraCegoVer} subset necessitated a thorough analysis of the classification errors. To that end, we conducted a detailed analysis of all FP and FN samples to identify the factors contributing to these outcomes. Below, we discuss the findings from this analysis.

\subsubsection{Analysis of False Positive Classifications}

\begin{figure*}[!t]
	\begin{tabular}{cccc}
    \centering
    \includegraphics[width=0.22\textwidth,height=0.15\textwidth]{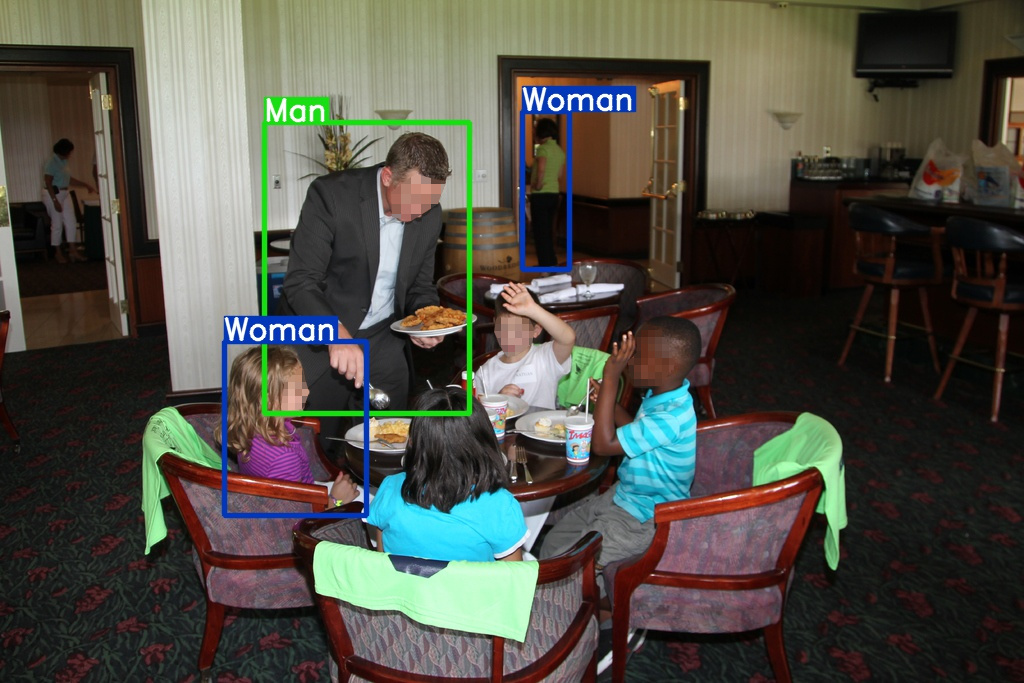} & 
    \includegraphics[width=0.22\textwidth,height=0.15\textwidth]{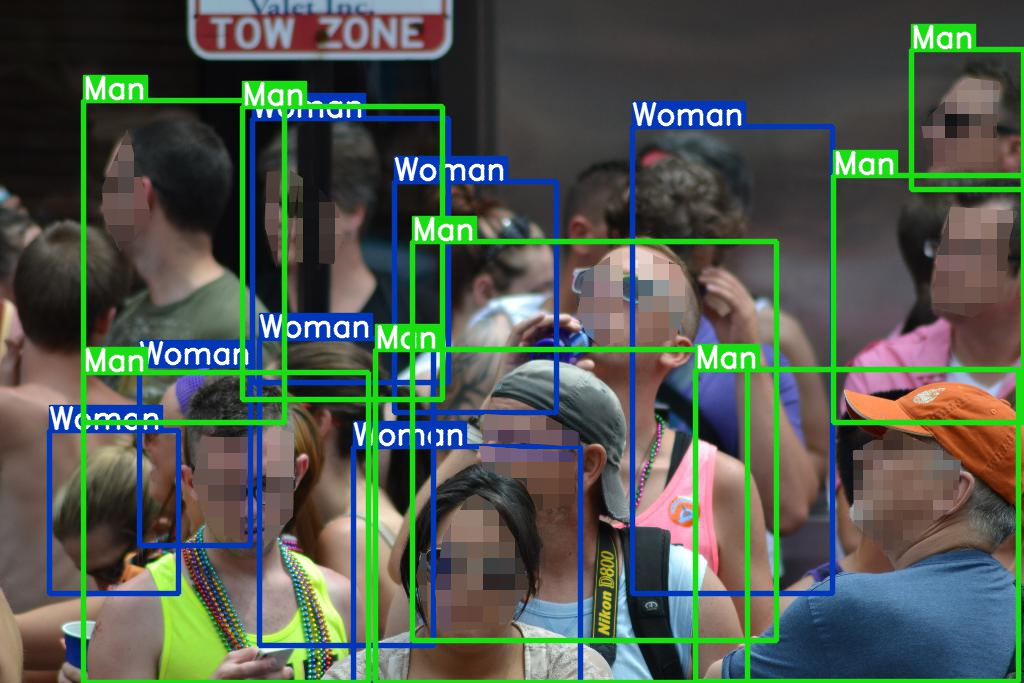} &   
    \includegraphics[width=0.22\textwidth,height=0.15\textwidth]
    {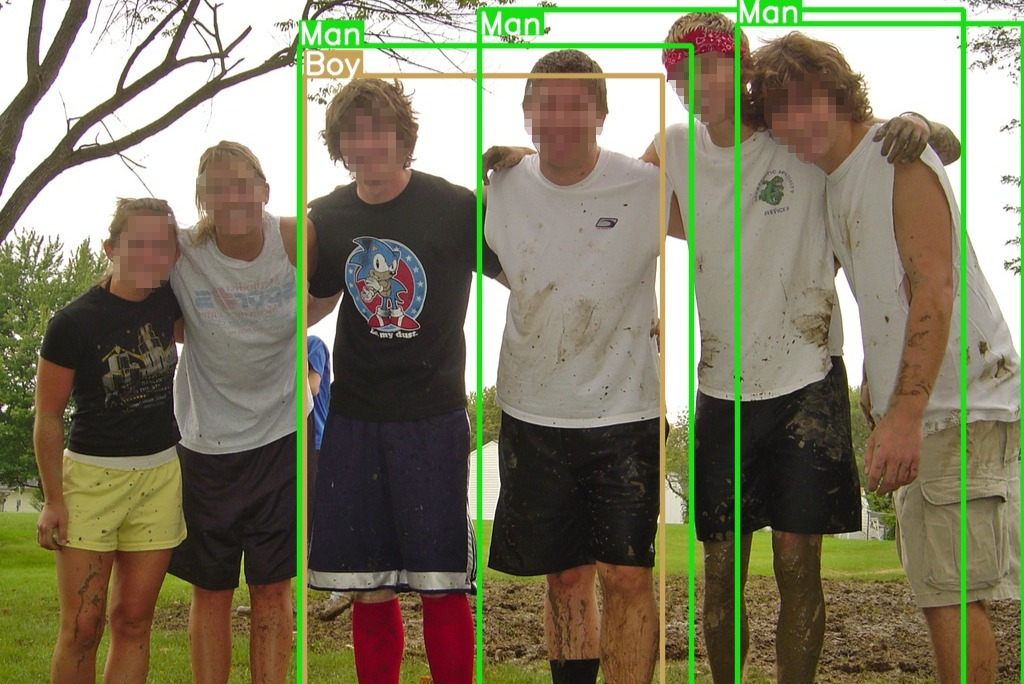} & 
    \includegraphics[width=0.22\textwidth,height=0.15\textwidth]{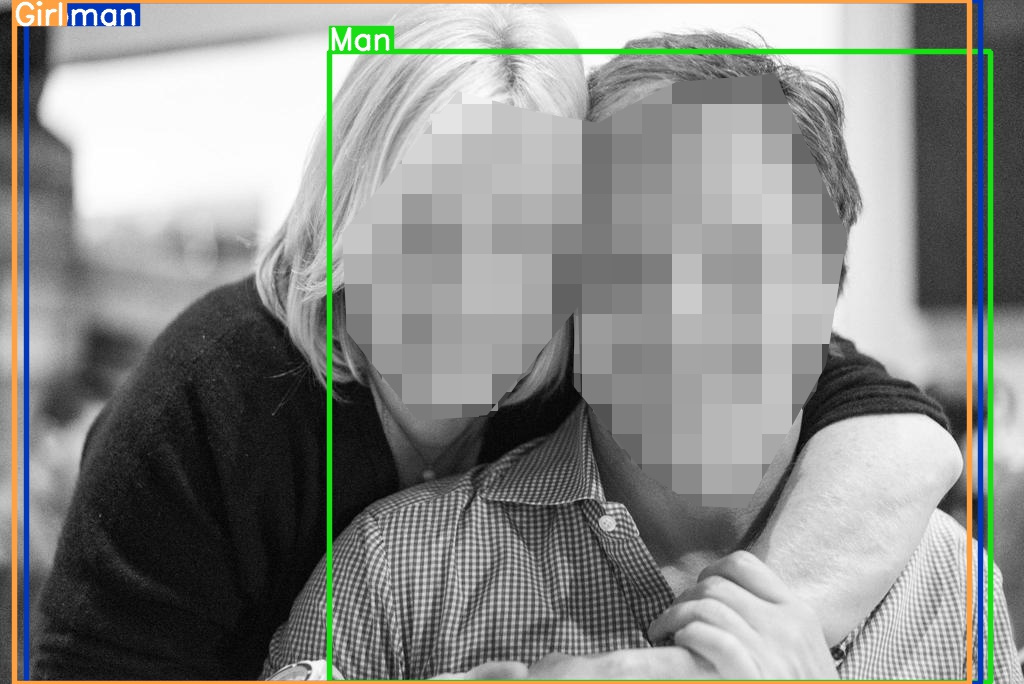} \\

    \includegraphics[width=0.22\textwidth,height=0.15\textwidth]{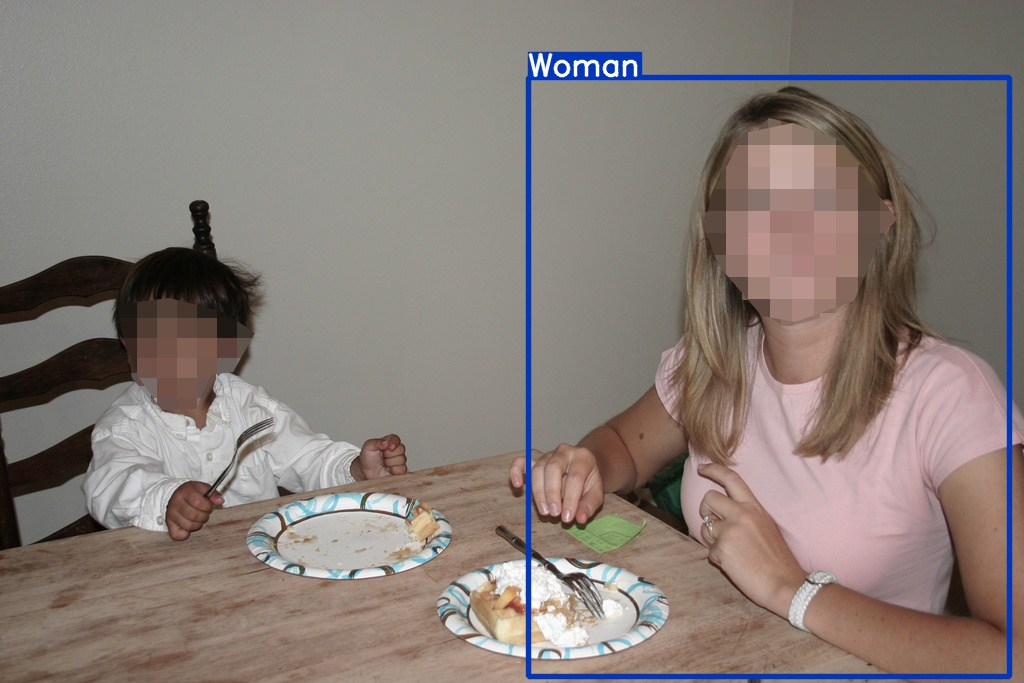} & 
    \includegraphics[width=0.22\textwidth,height=0.15\textwidth]{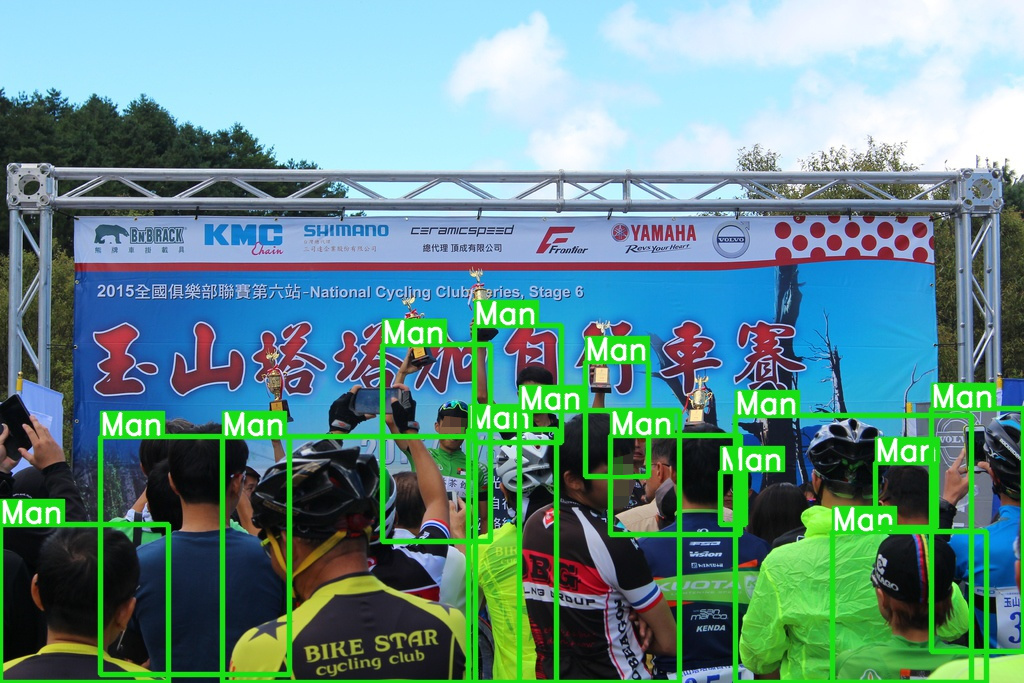} &   
    \includegraphics[width=0.22\textwidth,height=0.15\textwidth]
    {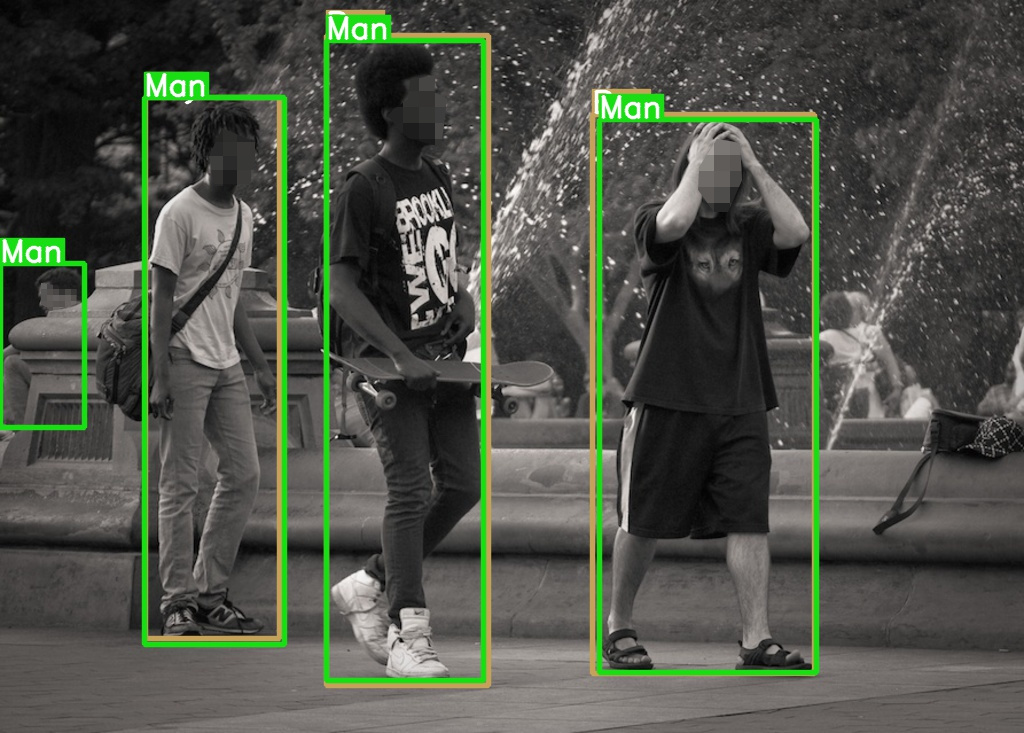} & 
    \includegraphics[width=0.22\textwidth,height=0.15\textwidth]{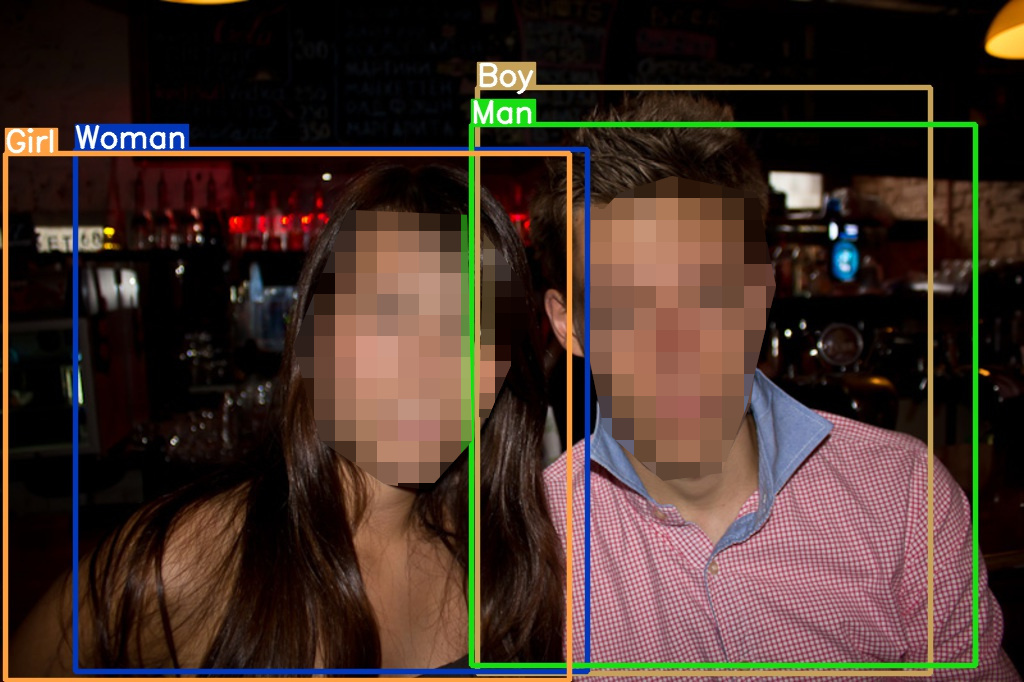} \\

        \includegraphics[width=0.22\textwidth,height=0.15\textwidth]{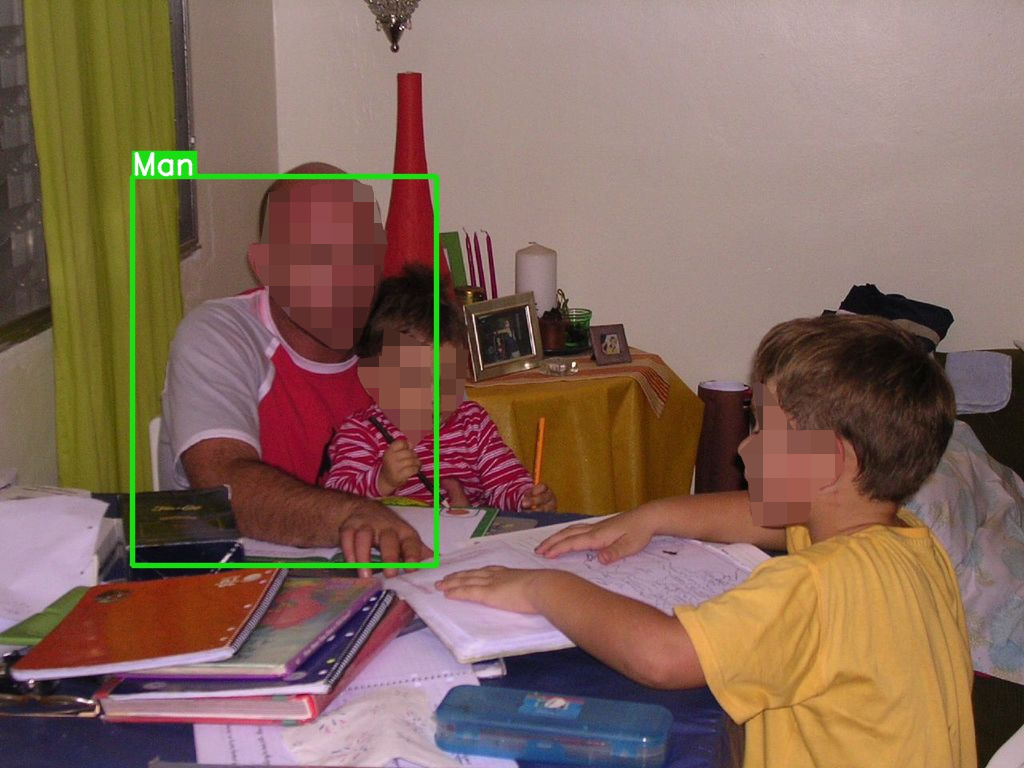} & 
    \includegraphics[width=0.22\textwidth,height=0.15\textwidth]{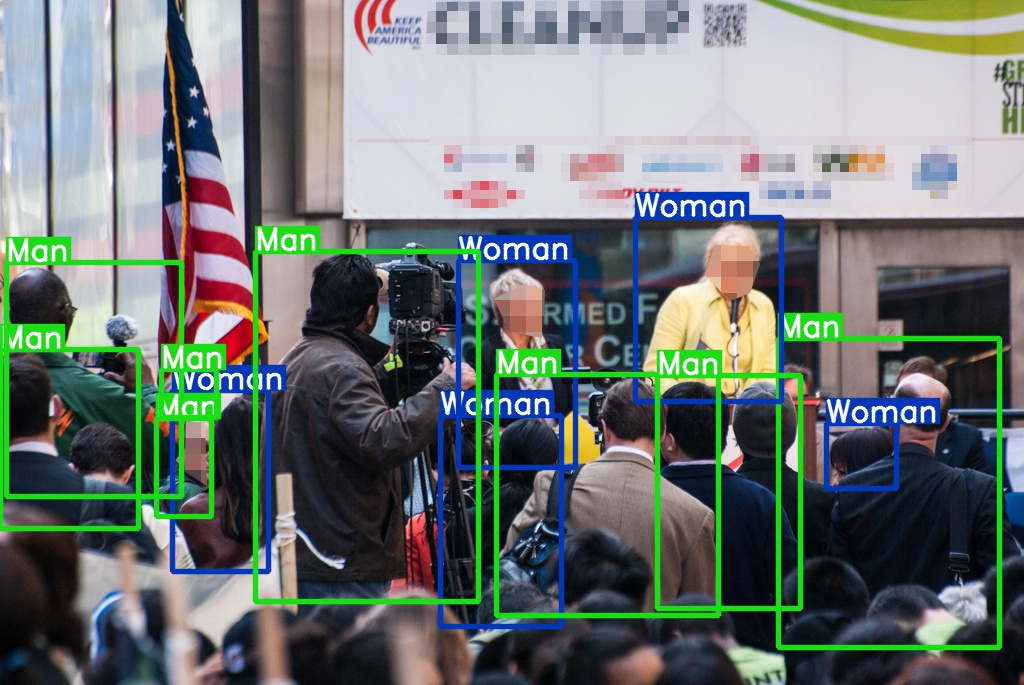} &   
    \includegraphics[width=0.22\textwidth,height=0.15\textwidth]
    {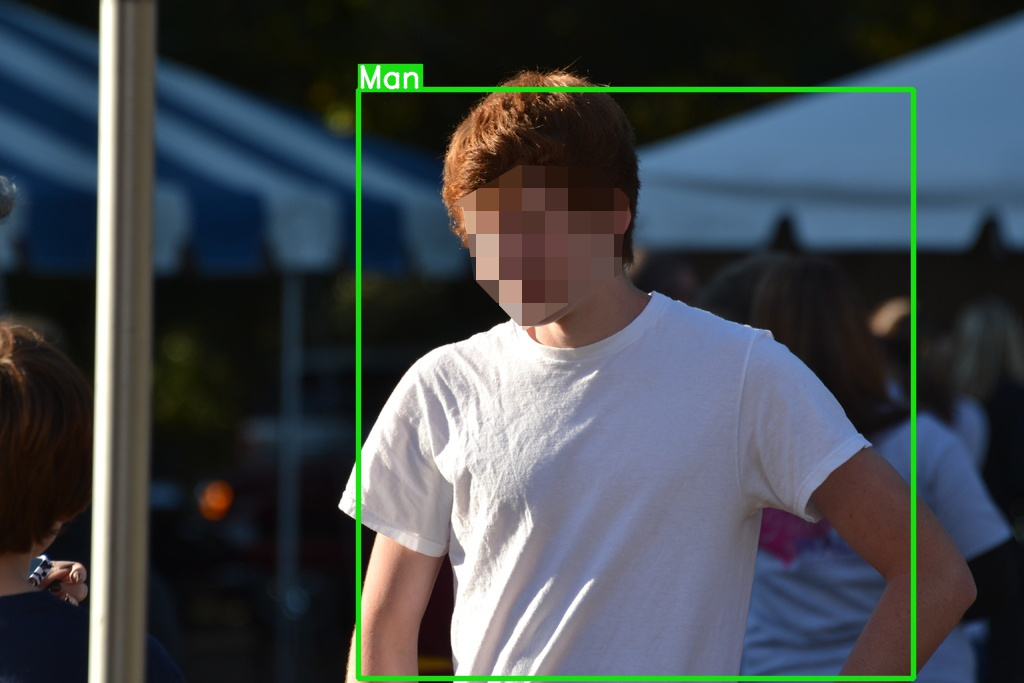} & 
    \includegraphics[width=0.22\textwidth,height=0.15\textwidth]{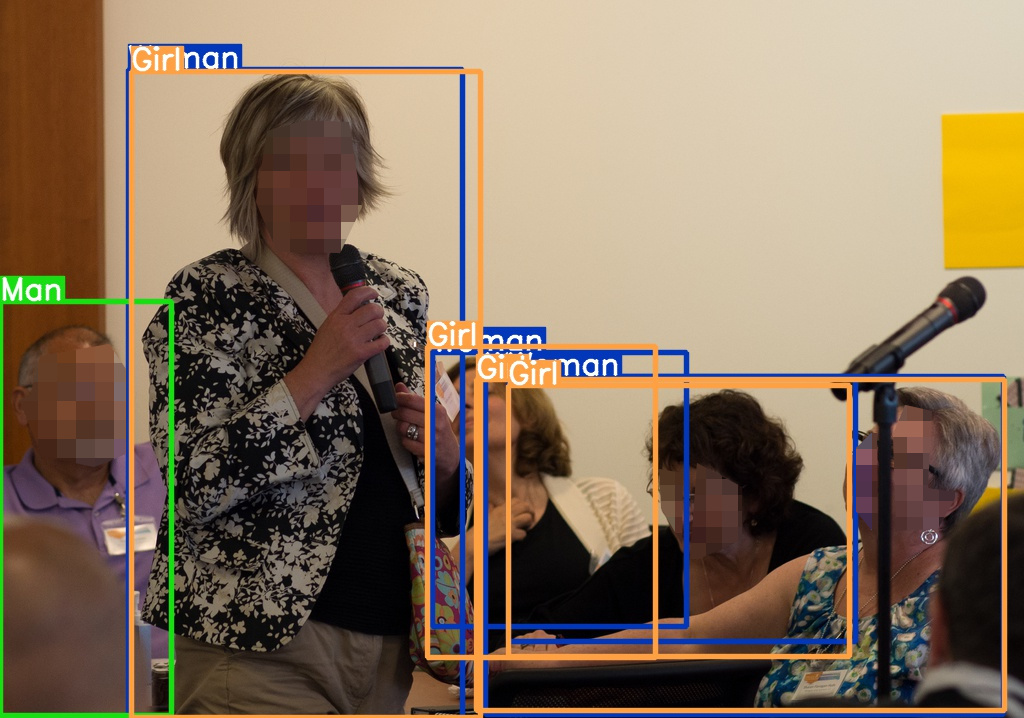} \\
    
    \footnotesize{(a) Mislabeled or missing annotations} & 
    \footnotesize{(b) Crowded scenes} & 
    \footnotesize{(c) Teenagers} & 
    \footnotesize{(d) Double annotations} \\
	\end{tabular}
	\caption{Representative examples of False Positives (FP) and False Negatives (FN) identified in the Open Images V7 subset. (a) Mislabeled or missing annotations. (b) Crowded scenes with numerous individuals frequently resulted in FP classifications. (c) Teenagers were challenging to classify due to their resemblance to adults, leading to FN errors. (d) Double annotations, where the same individual was labeled as both ``Boy'' and ``Man'' or ``Girl'' and ``Woman''. To protect privacy, the faces of all individuals in the images were pixelated. These examples highlight annotation inconsistencies and dataset complexities that influenced the model performance.}
    
    \Description{Representative examples of False Positives (FP) and False Negatives (FN) identified in the Open Images V7 subset. (a) Mislabeled or missing annotations. (b) Crowded scenes with numerous individuals frequently resulted in FP classifications. (c) Teenagers, were challenging to classify due to their resemblance to adults, leading to FN errors. (d) Double annotations, where the same individual was labeled as both ``Boy'' and ``Man'' or ``Girl'' and ``Woman''. To protect privacy, the faces of all individuals in the images were pixelated. These examples highlight annotation inconsistencies and dataset complexities that influenced the model performance.}
	\label{fig:oi_FP_FN}
\end{figure*}

We identified several key factors contributing to the misclassifications upon analyzing the FP classifications. A prominent issue was the presence of incorrect annotations within the Open Images V7 dataset. Specifically, we found several child images mislabeled with ``Man'' or ``Woman'' tags, leading to incorrect identification as non-child images during the annotation process. Additionally, a significant number of images with children were entirely missing the expected ``Boy'' or ``Girl'' labels, leaving them unannotated and contributing to errors. Both mentioned problems are illustrated in \autoref{fig:oi_FP_FN}~(a). This omission underscores the limitations of the dataset’s annotation process.

Moreover, some FP involved images containing unidentifiable children due to occlusions or lack of visible facial features. Despite their presence, these images were challenging to classify accurately based on the visual input provided. Another contributing factor was including cartoon or illustration images, which persisted despite efforts to filter them out during the subset creation. This highlights additional errors in the annotation process. Lastly, images depicting crowded scenes with numerous individuals were particularly prone to FP despite the absence of children. Such example is illustrated in \autoref{fig:oi_FP_FN}~(b). 

\subsubsection{Analysis of False Negative Classifications}

The analysis of FN classifications revealed several recurring challenges that impeded accurate classification. A notable difficulty was the presence of teenagers in the dataset. While annotated as ``Boy'' or ``Girl,'' teenagers often exhibit features that resemble adults, making them inherently challenging to classify even for human evaluators. This ambiguity was a common source of FN and is illustrated in \autoref{fig:oi_FP_FN}~(c).

Another critical issue was the prevalence of double annotations, where the same individual in an image was labeled both as ``Boy'' and ``Man'', or ``Girl'' and ``Woman''. We found it to be especially pronounced in cases involving ``Girl'' and ``Woman'' annotations, see \autoref{fig:oi_FP_FN}~(d). Such discrepancies underscore deeper concerns about potential biases --- particularly sexist undertones --- permeating the annotation process. Even more alarming, numerous images labeled as ``Girl'' were, in reality, adult women. This mislabeling not only highlights significant inconsistencies but also reveals a deeper systemic issue of bias embedded within the dataset's annotation framework.

\subsubsection{Discussion}
The lower performance on the Open Images V7 subset illustrates the challenges of relying on datasets with significant annotation inconsistencies and biases. Through the comprehensive analysis of all FP and FN samples, we identified critical limitations in the dataset's annotation process, including missing labels and systemic biases.

These findings emphasize the importance of rigorous dataset auditing and annotation refinement to improve the reliability of AI models. Moreover, the results underscore the necessity of prioritizing Recall to ensure all child images are accurately identified, aligning with the study's ethical objectives. While FPR remains a key indicator of misclassification, the ultimate priority remains the detection and removal of child images to safeguard their privacy and safety.

\subsection{Carbon Footprint}

One of the significant challenges in deploying deep learning models is the substantial computational resources required for training and inference. To assess the environmental implications of our experiments, we employed the CodeCarbon tool~\cite{benoit:2024} to monitor energy consumption during the inference phase. We emphasize this study did not encompass the training phase of models, a process widely recognized as the primary contributor to carbon emissions in machine learning workflows. Instead, our investigation was limited to performing inference using pre-trained models.

While computationally less intensive than training, the inference process still incurs energy costs. We tracked the power usage of our hardware configurations and estimated the carbon footprint associated with each model. Table~\ref{tab:energy_consumption} details energy consumption metrics for the models evaluated.

\begin{table*}[t]
    \centering
    \caption{Average energy consumption and equivalent CO$_2$ emissions (CO$_2$-eq) for each evaluated model, alongside the number of parameters, for inference over 100,000 images.}
    \label{tab:energy_consumption}
    \begin{tabular}{c l c c c}
        \toprule
        \textbf{Language Decoder} & \textbf{Model}  & \textbf{\# Parameters (B)} & \textbf{Energy (kWh)} & \textbf{CO2-eq (kg)} \\
        \toprule
        \multirow{6}{*}{Vicuna~\cite{Chiang:2023:Vicuna}} & llava-v1.5-7b~\cite{Liu:2024:CVPR:LLaVA1.5} & 7 &  5.759 & 0.566 \\
        & llava-v1.5-7b-lora~\cite{Liu:2024:CVPR:LLaVA1.5} & 7 & 7.587 &  0.746 \\
        & llava-v1.5-13b~\cite{Liu:2024:CVPR:LLaVA1.5} & 13 & 7.857 &  0.773\\
        & llava-v1.5-13b-lora~\cite{Liu:2024:CVPR:LLaVA1.5} & 13 & 6.318 & 0.621 \\
        & llava-v1.6-vicuna-7b~\cite{Liu:2024:LLaVA1.6-NeXT} & 7 & 11.656 &  1.146 \\
        & llava-v1.6-vicuna-13b~\cite{Liu:2024:LLaVA1.6-NeXT} & 13 & 16.088  &  1.582 \\
        \midrule
        Mistral~\cite{Jiang:2023:Mistral} & llava-v1.6-mistral-7b~\cite{Liu:2024:LLaVA1.6-NeXT} & 7 & 13.493  & 1.327 \\
        \midrule
        Hermes-Yi~\cite{NousHermesYi:2024} & llava-v1.6-34b~\cite{Liu:2024:LLaVA1.6-NeXT} & 34 & 34.948  &  3.427 \\
        \midrule
    \end{tabular}
\end{table*}

\subsection{A Low-Cost Method for Removing Images of Children}

The proposed pipeline leverages VLMs' remarkable generalization capabilities across a wide range of tasks, including the VQA task that plays a key role in our approach. This versatility allows for robust performance in addressing complex scenarios, which is crucial when processing images from noisy datasets like \textit{\#PraCegoVer}, LAION family, and others. Nevertheless, these models require expensive computational costs and a significant amount of time, which can hamper from a logistic and an environmental perspective.

Considering this, we view the development and usage of lightweight models, fine-tuned to be task-specific rather than generalist, as a promising approach for detecting and filtering images containing children. An alternative method, which could employ low-cost techniques, involves related tasks, such as age estimation. Age estimation typically consists of two stages: detecting the faces within an image and estimating the age of each identified face. Images including children could be effectively identified by analyzing the resulting age data across all samples in the dataset. The relative simplicity of the age estimation task, compared to VQA, provides several benefits, including reduced processing times and lower computational demands, which also leads to a decreased carbon footprint and environmental impact.

Our preliminary experiments used the age estimation model described in~\citet{macedo2018rcpd}. We observed an energy cost of 0.658 kWh and 0.065 kg equivalent CO2 emissions for 100,000 images, coming off as approximately 17.7 times lower than the best-performing model (llava-v1.6-vicuna-7b). Despite these improvements, the model performance was not sufficiently reliable for our specific use case, where avoiding false negatives is crucial and presents a significant challenge, with a $0.773$ balanced accuracy score and a $0.858$ recall score. Given this, we hope that, by raising this concern, future work in the field will be steered towards creating lightweight models that are both high performing and efficient for age estimation tasks, addressing a balance between accuracy and resource optimization.
\section{Limitations}\label{sec:limitations}

Our evaluation settings were limited by the nature of the datasets used in the experiments. Specifically, the subset derived from \textit{\#PraCegoVer} was meticulously curated, leveraging the image metadata to verify the presence or absence of children. However, this selection process produced a relatively small subset of $1,364$ samples, restricting the scope of our evaluation. In contrast, the subset from Open Images comprised $100,000$ samples with age-related annotations. However, the dataset exhibited significant inconsistencies, including missing labels for children and numerous labeling conflicts where the same individual was simultaneously assigned both adult and child-related labels. 

Another significant limitation is the absence of bias assessment in our study. Given our objective of removing images of children from large-scale datasets to uphold children’s right to data safety, it is essential to evaluate potential performance disparities across protected attributes such as ethnicity and gender. While the datasets used in this study do not provide sufficient information to conduct a comprehensive analysis, future research should prioritize incorporating bias assessments.

Lastly, the proposed methodology does not guarantee the complete removal of images of children from datasets. Therefore, applying our methodology does not grant unrestricted permission to utilize a given dataset, assuming all images of children have been removed. While our work represents a step toward this goal, the identified limitations highlight that we are still far from fully achieving it.

\section{Conclusion}\label{sec:conclusion}

This paper underscores the critical need to address ethical concerns associated with using child images in publicly available datasets. By developing and evaluating a visual-language pipeline for detecting and filtering child images, we provide an effective tool to mitigate potential harms arising from including such images in training data. The experimental results demonstrate the pipeline’s ability to achieve high Recall rates while maintaining low FPR, particularly with carefully designed prompts and model configurations.

However, our work also highlights significant challenges and limitations, including dataset inconsistencies, annotation biases, and the absence of comprehensive bias assessments. These issues underscore the need for rigorous dataset auditing and the development of more robust methodologies to address ethical and technical challenges.

While our methodology represents a step toward more ethical AI practices, it does not guarantee the complete removal of child images, nor does it eliminate the potential for misuse. The ethical risks associated with re-purposing this technology and its potential impact on downstream tasks call for ongoing vigilance, stricter safeguards, and transparent usage policies.

Future research could focus on conducting additional experiments related to age estimation, particularly leveraging zero-shot techniques (i.e., methods that do not use child data for training or fine-tuning). Assessing the impact of removing children's images on downstream tasks will be a critical direction, ensuring that ethical standards are upheld while supporting AI applications involving representations of children. Furthermore, thorough bias assessments need to be prioritized to identify and mitigate disparities, particularly in addressing sexist undertones in dataset annotations and model predictions.
\section*{Ethical Considerations Statement}\label{ethical_considerations}

The objective of our work is intrinsically aligned with mitigating a pressing and growing concern in the machine learning community: the indiscriminate use of images of children in datasets, which can lead to significant downstream harms when models trained on such images are deployed. Furthermore, during the development of this work, we addressed the following ethical concerns.

\paragraph{Anonymizing images} To ensure the privacy of individuals depicted in the datasets, whether children or adults, we implement an anonymization procedure by pixelating their faces in all images illustrating this article. This approach ensures no direct exposure of people's identities while allowing qualitative assessment of our results.

\paragraph{Curating a reliable validation subset} A prevalent challenge in age-related tasks is the reliance on the annotators' subjective perceptions of age, which can be influenced by factors such as experience, cultural background, and personal biases. To address this, we used the comprehensive metadata and captions available in the \textit{\#PraCegoVer} dataset to curate a validation subset with higher reliability when distinguishing adults from children.

\section*{Adverse Impacts Statement}
The following is a list of potential adverse impacts that we anticipate as a result of this work.

\paragraph{Misuse of child removal pipeline.} Malicious actors could repurpose our pipeline to locate and isolate images of children from any collection of images for unknown or unethical purposes, potentially facilitating the exploitation or unauthorized use of such images. Our pipeline relies on a VLM that automatically identifies images of children through prompts, making it viable to replicate with minimal programming expertise. This highlights the importance of implementing robust safeguards and usage policies for both datasets and algorithms handling data related to children.

\paragraph{Leveraging pre-trained visual-language models} This work experiments VLMs originally trained on a large-scale dataset containing images, captions, and other data associated with children. This contradicts our long-term objective of discouraging the use of data from children unless it is collected following legal and ethical guidelines. \new{We do not encourage or endorse training models on such data.} Furthermore, as others may replicate and build upon our work, it inadvertently incentivizes the ongoing use of models trained on such data. \new{While we leverage existing technology, our work applies it in a novel way to draw attention to an urgent issue: the widespread presence of children's images in large-scale scraped datasets. Importantly, we do not advocate for the systematic use of these models to remove children's images. Rather, this paper is a call to action for the community to prioritize child protection and to critically reflect on the ethical implications of training and deploying such models.}

\paragraph{Unknown impact on downstream tasks.} Large-scale datasets serve as the foundation for training large, multipurpose models that underpin various downstream tasks. Removing images of children from such datasets is likely to have a negative impact on downstream tasks, including the ones for which safe and legal handling of images of children is viable. Therefore, future work should focus on designing experiments to assess the impact of removing such images from training data, ensuring that these actions do not inadvertently hinder the performance of derived models on applications that rely on accurate representations of children.

\section*{Acknowledgments}

This work is partially funded by FAPESP 2023/12086-9, PIND/ FAEPEX/UNICAMP 2597/23, and the Serrapilheira Institute R-2011-37776. C.~Caetano (2024/01210-3), G.~O.~dos~Santos (2024/07969-1), C.~Petrucci (2024/09375-1), A.~Barros (2024/09372-2), L.~S.~F.~Ribeiro (2022/14690-8), and S.~Avila (2020/09838-0, 2013/08293-7) are also funded by FAPESP. S.~Avila is also funded by H.IAAC 01245.003479/ 2024-10 and CNPq 316489/2023-9.

\bibliographystyle{ACM-Reference-Format}
\bibliography{bibliography}

\end{document}